\documentclass{article}

\usepackage{microtype}
\usepackage{graphicx}
\usepackage{subfigure}
\usepackage{booktabs} 

\usepackage{hyperref}
\usepackage{amsmath,amssymb}
\usepackage{multirow}
\usepackage{dsfont}
\usepackage{enumitem}
\usepackage{CJKutf8}

\usepackage{siunitx}

\DeclareMathOperator*{\argmax}{arg\,max}

\newcommand*{\round}[1]{\num[round-mode=places,round-precision=1]{#1}}

\def\smup{sMup}
\def\smlp{sMlp}
\def\hmup{hMup}
\def\hmlp{hMlp}

\usepackage[accepted]{icml2019}

\icmltitlerunning{
Mixture Models for Diverse Machine Translation: Tricks of the Trade}

\begin{document}

\twocolumn[
\icmltitle{Mixture Models for Diverse Machine Translation: Tricks of the Trade}



\icmlsetsymbol{equal}{*}

\begin{icmlauthorlist}
\icmlauthor{Tianxiao Shen}{equal,mit}
\icmlauthor{Myle Ott}{equal,fb}
\icmlauthor{Michael Auli}{fb}
\icmlauthor{Marc'Aurelio Ranzato}{fb}
\end{icmlauthorlist}

\icmlaffiliation{mit}{MIT CSAIL}
\icmlaffiliation{fb}{Facebook AI Research}

\icmlcorrespondingauthor{Tianxiao Shen}{tianxiao@csail.mit.edu}

\icmlkeywords{Machine Learning, ICML}

\vskip 0.3in
]



\printAffiliationsAndNotice{\icmlEqualContribution} 

\begin{abstract}
Mixture models trained via EM are among the simplest, most widely used and well understood latent variable models in the machine learning literature. Surprisingly, these models have been hardly explored in text generation applications such as machine translation.
In principle, they provide a latent variable to control generation and produce a diverse set of hypotheses. 
In practice, however, mixture models are prone to degeneracies---often only one component gets trained or the latent variable is simply ignored.
We find that disabling dropout noise in responsibility computation is critical to successful training. In addition, the design choices of parameterization, prior distribution, hard versus soft EM and online versus offline assignment can dramatically affect model performance.
We develop an evaluation protocol to assess both quality and diversity of generations against multiple references, and provide an extensive empirical study of several mixture model variants.
Our analysis shows that certain types of mixture models are more robust and offer the best trade-off between translation quality and diversity compared to variational models and diverse decoding approaches.\footnote{Code to reproduce the results in this paper is available at \url{https://github.com/pytorch/fairseq}}
\end{abstract}

\section{Introduction} \label{sec:intro}
Machine translation (MT) is a challenging task not only because of the large and structured output space, but also because it is inherently a one-to-many mapping. 
There are often many plausible and semantically equivalent translations due to information asymmetry between different languages, e.g., translating from a language without grammatical gender to a language that has grammatical gender leads to two valid translation options, as well as different translation styles such as formal/informal, literal/not literal, etc.
This raises the question of how to model such {\em multi-modal} output distributions and how to evaluate these models. 

Our first contribution is a better evaluation protocol that uses multiple references during evaluation to measure both the {\em quality} of translation and {\em diversity} of a generated hypothesis set.
The second contribution of this paper is an in-depth empirical analysis of mixture models for machine translation, although we conjecture that the findings are general and might apply to other text generation tasks, such as dialogue, summarization, image captioning, etc.

Conditional mixture models, also known as mixture of experts (MoE) \cite{moe91}, are in principle well suited to generating diverse hypotheses which can be achieved through different mixture components.
However, they have been largely overlooked in favor of models with richer latent structure~\cite{zhang2016variational,kaiser18}.
There has been some previous work on mixture models for sequence to sequence learning~\cite{largemoe17,he2018sequence}, but these did not evaluate generations in terms of both quality and diversity, or they focus on a particular model variant.
There is a lack of consensus whether mixture models are competitive with more complex models that rely on approximate Bayesian inference, whether they are plagued by the same ``posterior collapse" degeneracy as variational models~\cite{bowman15}, how model configurations affect performance and which one works best in practice. 

This work considers all the major design choices involved in the construction of mixture models, including hard versus soft EM training, different parameterizations of mixture components, the choice of conditional prior, update frequency of responsibilities (also called membership weights), and how regularization noise is injected. 
We experiment on the large scale WMT English to German benchmark with a state-of-the-art model architecture and the results demonstrate intricate dependencies between these design choices. They also reveal that some ingredients are key to successful training of mixture models.

First, we show that mixture models are prone to degeneracies when trained with dropout noise, but that this can be mitigated by turning off dropout in the computation of responsibilities.
The key to the specialization of experts is to make consistent use of them, and even a small amount of regularization noise can hamper that.
Second, hard mixtures yield more diverse generations than soft mixtures, similar to how K-Means tends to find centroids that are farther apart from each other compared to the means found by a mixture of Gaussians~\cite{kearns1998information}.
Third, employing a uniform prior encourages {\em all} mixture components to produce good translations for any input source sentence, which is highly desirable.
Finally, using independently parameterized mixture components provides greater diversifying capacity than shared parameters; but if responsibilities are refreshed online, independent parameterization is prone to a degeneracy where only a single component is trained because of the ``rich gets richer" effect.
Conversely, the combination of shared parameters and offline responsibility assignment may lead to another degeneracy, in which the mixture components fail to specialize and behave the same.

We extend our evaluation to three WMT benchmark datasets for which test sets with multiple human references are available.
We demonstrate
that mixture models, when successfully trained,
consistently outperform variational NMT \cite{zhang2016variational} and diverse decoding algorithms such as diverse beam search~\citep{jiweili17,vijayakumar2018diverse} and biased sampling~\citep{graves13,fan2018hierarchical}.
Our qualitative analysis shows that different mixture components can capture consistent translation styles across examples, enabling users to control generations in an interpretable and semantically meaningful way.


\section{Related Work}
Prior studies have investigated the prediction uncertainty in machine translation.
\citet{hyter} and \citet{galley2015deltableu} introduced new metrics to address uncertainty at evaluation time.
\citet{ott_icml18} inspected the sources of uncertainty and proposed tools to check fitting between the model and the data distributions. 
They also observed that modern conditional auto-regressive NMT models can only capture uncertainty to a limited extent, and they tend to oversmooth probability mass over the hypothesis space.

Recent work has explored latent variable modeling for machine translation.
\citet{zhang2016variational} leveraged variational inference~\cite{kingma2014auto, bowman15} to augment an NMT system with a single Gaussian latent variable.
This work was extended by~\citet{schulz2018stochastic}, who considered a sequence of latent Gaussian variables to represent each target word.
\citet{kaiser18} proposed a similar model, but with groups of discrete multinomial latent variables.
In their qualitative analysis, \citet{kaiser18} showed that the latent codes do affect the output predictions in interesting ways, but their focus was on speeding up regular decoding rather than producing a diverse set of hypotheses.
None of these works analyzed and quantified diversity introduced by such latent variables.

The most relevant work is by~\citet{he2018sequence}, who propose to use a soft mixture model with uniform prior for diverse machine translation.
However, they did not evaluate on datasets with multiple references, nor did they analyze the full spectrum of design choices for building mixture models.
Moreover, they used weaker base models and did not compare to variational NMT or diverse decoding baselines, which makes their empirical analysis less conclusive.
We provide a comprehensive study and shed light on the different behaviors of mixture models in a variety of settings.


Besides machine translation, there is work on latent variables for dialogue generation~\citep{serban2017hierarchical,cao2017latent,wen2017latent} and image captioning~\citep{wang2017diverse,dai2017towards}.
The proposed mixture model departs from these VAE or GAN-based approaches and importantly, is much simpler. It could also be applied to other text generation tasks as well.

\section{Mixture Models for Diverse MT}
\label{sec:models}

A standard neural machine translation (NMT) model has an encoder-decoder structure.
The encoder maps a source sentence $x$ to a sequence of hidden states, which are then fed to the decoder to generate an output sentence one word at a time.
At each time step, the decoder additionally conditions its output on the previous outputs, resulting in an auto-regressive factorization $p(y|x;\theta)=\prod_{t=1}^T p(y_t|y_{1:t-1},x;\theta)$, where $(y_1,\cdots,y_T)$ are the words that compose a target sentence $y$.

However, the machine translation task has inherent uncertainty, due to the existence of multiple valid translations $y$ for a given source sentence $x$.
With the auto-regressive factorization all uncertainty is represented in the decoder output distribution, making it difficult to search for multiple modes of $p(y|x;\theta)$.
Indeed, widely used decoding algorithms such as beam search typically produce hypotheses of low diversity with only minor differences in the suffix~\cite{ott_icml18}.

Mixture models provide an alternative approach to modeling uncertainty and generating diverse translations.
While these models have primarily been explored as a means of increasing model capacity~\cite{moe91,largemoe17}, they are also a natural way of modeling different translation styles~\cite{he2018sequence}.

Formally, given a source sentence $x$ and reference translation $y$, a mixture model introduces a multinomial latent variable $z\in \{1,\cdots,K\}$, and decomposes the marginal likelihood as:
\begin{align}
    p(y|x;\theta)=\sum_{z=1}^K p(y,z|x;\theta)
    = \sum_{z=1}^K p(z|x;\theta)p(y|z,x;\theta)
\end{align}
where the prior $p(z|x;\theta)$ and likelihood $p(y|z,x;\theta)$ are learned functions parameterized by $\theta$.
Each value of $z$ represents an \emph{expert}, and the posterior probability:
\begin{align}
    p(z|x,y;\theta)=\frac{p(z|x;\theta)p(y|z,x;\theta)}{\sum_{z'} p(z'|x;\theta)p(y|z',x;\theta)}
\end{align}
can be viewed as the \emph{responsibility} each expert takes for explaining an observation $(x,y)$.

\paragraph{Training}
Given a training set $\{(x^{(i)}, y^{(i)})\}_{i=1}^{N}$, we want to find $\theta$ that maximizes the log likelihood. To this end, for each training example we compute the gradient:
\begin{align}\label{eq:grads}
    \nabla_\theta \log p(y^{(i)}|x^{(i)};\theta)=
    \sum_z ~&p(z|x^{(i)},y^{(i)};\theta)~\cdot\nonumber\\
    &\nabla_\theta\log p(y^{(i)},z|x^{(i)};\theta)
\end{align}
and train the model with the EM algorithm~\citep{dempster1977maximum} by iteratively applying the following two steps:
\begin{description}[topsep=0pt]
    \item [E-step:] estimate the responsibilities of each expert $r_z^{(i)}\leftarrow p(z|x^{(i)},y^{(i)};\theta)$ using the current parameters $\theta$;
    \item [M-step:] update $\theta$ through each expert with gradients $\nabla_\theta\log p(y^{(i)},z|x^{(i)};\theta)$ weighted by their responsibilities $r_z^{(i)}$.
\end{description}
There are several ways to apply the E and M steps across training examples~\citep{neal1998view}; \S\ref{subsec:schedule} will review those considered in this work.

\paragraph{Decoding}
All the decoding strategies for $p(y|x;\theta)$ of a baseline model can be applied equally to $p(y|z,x;\theta)$ in a mixture model. We adopt the most straightforward one: generating $K$ hypotheses by first enumerating $z$ and then greedily decoding $\hat y_t=\argmax_y p(y|\hat y_{1:t-1},z,x;\theta)$.
Notably, this decoding procedure is efficient and easily parallelizable.

\paragraph{Degeneracies}
Unfortunately, na\"ive implementations of mixture models for text generation are prone to two major types of degeneracies:
\begin{description}[topsep=0pt]
    \item [\textbf{D1:}] Only one component gets trained~\cite{eigen14,largemoe17} because of the ``rich gets richer" effect whereby, once a component is slightly better than others, it is always picked while the other components starve and are eventually never used.
    \item [\textbf{D2:}] The latent variable is ignored, similar to the collapse of variational auto-encoders where the posterior is always equal to the prior~\cite{bowman15}.
\end{description}
In both cases, the model operates like a baseline model without any benefit from the latent variable.
In practice, however, the chance of these degeneracies is heavily affected by a number of design decisions, which we describe in the following subsections.

\subsection{Model Variants}
\label{subsec:model_variants}

Ideally, we would like different experts to specialize on different translation styles, so they can generate diverse hypotheses. Moreover, we want all of them to work well with any source sentence, so they will produce high quality translations.
To this end, we explore several variants of the canonical mixture model, which vary by whether they use hard (h) or soft (s) assignments of responsibilities, and whether they use a learned prior (lp) or uniform prior (up).

The specialization of experts implies that the responsibility $p(z|x,y;\theta)$ for explaining a particular translation $y$ should be large for only one $z$, i.e., only one element in the sum $\sum_z p(y,z|x;\theta)$ dominates.
To encourage this, a \emph{hard mixture model} directly optimizes for $\max_z p(y,z|x;\theta)$ by assigning full responsibility for each training example to the expert with the largest joint probability.
Training proceeds via hard-EM, where the M-step remains unchanged and the E-step becomes:
\begin{description}[topsep=0pt]
\item [E-step (hard):] estimate the responsibilities of each expert $r_z^{(i)}\leftarrow \mathds{1}[z=\argmax_{z'} p(y^{(i)},z'|x^{(i)};\theta)]$ using the current parameters $\theta$.
\end{description}
This can also be seen as maximizing the marginal likelihood $p(y|x;\theta)$ while minimizing the gap between the sum and the max, thus finding a balance between the two terms~\citep{kearns1998information}. 

To encourage all experts to generate good hypotheses for any source sentence, we may set the prior $p(z|x;\theta)$ to be uniform.
This can prevent the model from collapsing into only one working expert with extreme $p(z|x;\theta)$ value.
This also aligns with our simple decoding strategy that generates a single hypothesis from each expert.

The choices of soft (s) versus hard (h) mixture model (M) and learned prior (lp) versus uniform prior (up) give us four model variants with different loss functions:
\begin{align}\label{eq:loss}
    &\mathcal L_{\text{sMlp}}(\theta)=\mathbb E_{(x,y)\sim\text{data}}\left[-\log\sum_z p(z|x;\theta)p(y|z,x;\theta)\right] \nonumber\\
    &\mathcal L_{\text{sMup}}(\theta)=\mathbb E_{(x,y)\sim\text{data}}\left[-\log\sum_z p(y|z,x;\theta)\right] \nonumber\\
    &\mathcal L_{\text{hMlp}}(\theta)=\mathbb E_{(x,y)\sim\text{data}}\left[\min_z-\log p(z|x;\theta)p(y|z,x;\theta)\right] \nonumber\\
    &\mathcal L_{\text{hMup}}(\theta)=\mathbb E_{(x,y)\sim\text{data}}\left[\min_z-\log p(y|z,x;\theta)\right]
\end{align}
where the constant $\log K$ term is omitted for models with a uniform prior.

Among these variants, the hMup objective is also known as multiple choice learning (MCL) for an ensemble of learners where the oracle loss is minimized~\citep{guzman2012multiple}, and
\citet{he2018sequence} has considered the sMup objective to train a sequence-to-sequence mixture model.

\subsection{Parameterization}
\label{subsec:parameterization}

Another important design decision with mixture models is the degree of parameter sharing between experts.
Using \emph{independently} parameterized experts provides them with additional capacity to differentiate from one another, but may exacerbate overfitting since the number of parameters increases linearly with the number of experts.
On the other hand, \emph{sharing} parameters among experts may help mitigate degeneracy D1, whereby low quality experts are neglected and eventually ``die'' during training, since by sharing parameters even unpopular experts receive some gradients.

We test different model variants using both independent and shared parameters.
With independent parameterization, each expert has a different decoder network.
With shared parameterization, experts use the same decoder network but the beginning-of-sentence token at the start of the target sequence is replaced with an embedded representation of the latent variable.
This requires a negligible increase in parameters over the baseline model.

\subsection{Training Schedule}
\label{subsec:schedule}

We consider two schedules for alternating between the E-step and M-step during training: \emph{online} and \emph{offline}.
In the online EM algorithm, we minimize the loss via stochastic gradient descent, effectively interleaving the E-step and M-step for each mini-batch~\cite{lee16}.
In contrast, the offline EM algorithm performs the E-step for all training examples, trains each expert to convergence with the resulting responsibilities, and repeats.
In practice, for offline training we perform the M-step for only a single epoch, rather than to convergence, before re-estimating the responsibilities.



\subsection{Regularization}
\label{subsec:regularization}

Deep neural networks with a large amount of parameters are prone to overfitting, and regularization via dropout is usually adopted to achieving good generalization performance.
However, the key to make experts diversify as training progresses is to make consistent use of them, and we find that even a small amount of regularization noise in the computation of responsibilities (E-step) can hamper that.
Indeed, we show in \textsection\ref{subsec:moe_tricks} that na\"ive use of dropout causes mixture models to ignore the latent variable, but this degeneracy is mitigated by disabling dropout in the E-step.
We explore this phenomenon in more detail in Appendix~\ref{app:dropout}.
\section{Metrics}

\begin{figure}[t]
  \begin{center}
    \includegraphics[width=\columnwidth]{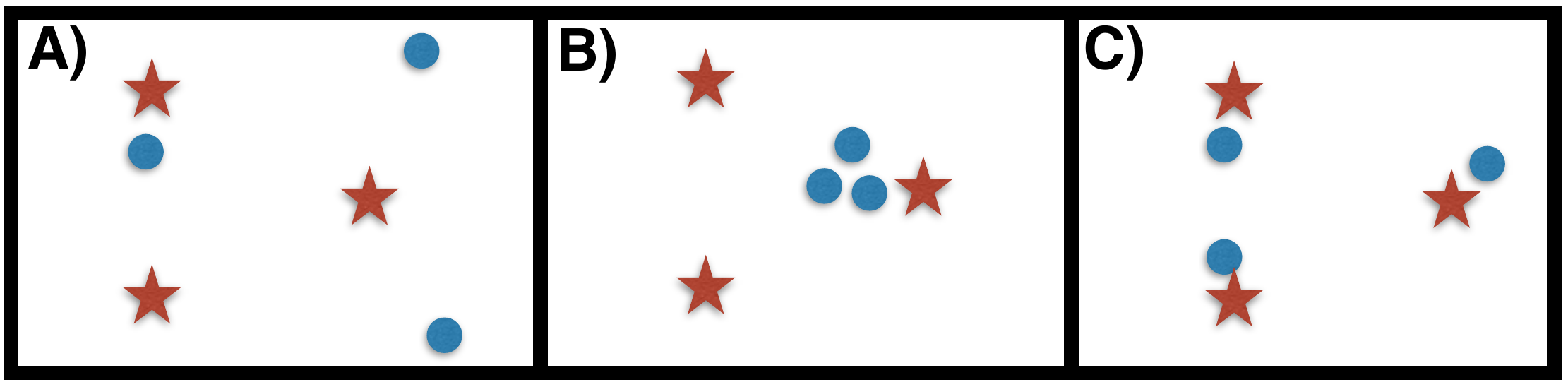}
  \end{center}
\caption{\small Toy illustration of how the metrics behave in various scenarios. Red stars represent human references, while blue dots represent system hypotheses.
{\bf A):} When there are poor hypotheses in the set, the BLEU is low  
and Pairwise-BLEU is low; this is the typical scenario of sampling hypotheses as well as degeneracy D1 described in  \textsection\ref{sec:models}.
{\bf B):} When hypotheses cluster closely together and in vicinity of a reference, the BLEU is high 
and Pairwise-BLEU is high; this is the typical scenario of beam search hypotheses and also degeneracy D2.
{\bf C):} When hypotheses match references (ideal case), the BLEU is high and Pairwise-BLEU matches human Pairwise-BLEU.}
\label{fig:toy_metrics}
\end{figure}

In this section, we describe the metrics we use to quantitatively assess the quality and diversity of a set of translation hypotheses. We use BLEU~\citep{papineni2002bleu} based on word n-gram matching to measure corpus similarity.

Suppose $\{y^1,\cdots,y^M\}$ are $M$ reference translations of a source sentence $x$, and $\{\hat y^1,\cdots,\hat y^K\}$ are $K$ hypotheses.
Let BLEU$\{([r_1,\cdots,r_n], h)\}_{x\in \text{data}}$ denote the corpus-level BLEU for all pairs where $h$ is a hypothesis and $[r_1,\cdots,r_n]$ is its reference list. Let $[n]$ denote the set of $\{1,\cdots,n\}$, and $[y^{-i}]$ denote $[y^1,\cdots,y^{i-1},y^{i+1},\cdots,y^M]$.
We compute the following two metrics\footnote{See Appendix~\ref{app:metric} for another diversity metric in terms of reference coverage.}~\citep{ott_icml18}:
\begin{itemize}[topsep=0pt]
\setlength\itemsep{1pt}

\item\textbf{Pairwise-BLEU:} To measure similarity among the hypotheses, we compare them with each other and compute BLEU$\{([\hat y^j], \hat y^k)\}_{x\in\text{data},j\in[K],k\in[K],j\ne k}$.\footnote{Several text GAN papers use Self-BLEU to evaluate diversity of unconditional generation~\citep{yu2017seqgan}, where each generated sentence is regarded as hypothesis against all other sentences as its reference list, i.e. BLEU$\{([\hat y^{-k}], \hat y^k)\}_{x\in\text{data},k\in[K]}$. For machine translation, Pairwise-BLEU offers a more precise evaluation of diversity compared to Self-BLEU. For example, if a source sentence has two valid translations $T_1\ne T_2$, system 1 provides 4 hypotheses $H_1=H_2=H_3=H_4=T_1$, and system 2 provides $H_1=H_2=T_1, H_3=H_4=T_2$, then their Self-BLEU are both 100; whereas Pairwise-BLEU for system 1 is 100 and for system 2 it is not, indicating that the latter is more diverse and more desirable (while they both have perfect quality).}
The more diverse the hypothesis set, the lower the Pairwise-BLEU.
Ideally, we would like a model with Pairwise-BLEU matching human Pairwise-BLEU.

\item \textbf{BLEU:} We calculate human BLEU in a leave-one-out manner by computing BLEU$\{([y^{-m}],y^m)\}_{x\in\text{data}}$ for $m\in[M]$ and then averaging the $M$ scores.
We also use $M-1$ references when computing system BLEU, to be comparable with human scores, i.e. average BLEU$\{([y^{-m}],\hat y^k)\}_{x\in\text{data},k\in[K]}$ for $m\in[M]$. This measures the {\em overall quality} of a hypothesis set. If this metric scores low, it implies that some generated hypotheses have poor quality.


    
\end{itemize}

Figure~\ref{fig:toy_metrics} illustrates how these metrics behave in different situations.
The degeneracies outlined in \textsection\ref{sec:models} can be easily identified with these metrics:
The first degeneracy is a situation where a single expert is responsible for all inputs (D1).
This case can be identified when we measure both very low Pairwise-BLEU as well as very low BLEU, i.e., all but one latent value produce very bad generations.
The second degeneracy happens when the latent variable is ignored and all experts behave almost identically (D2).
This is evident when we observe good BLEU but extremely high Pairwise-BLEU (close to 100), i.e., when all latent values give good yet very similar outputs.

\section{Experiments}
\label{sec:experiments}

\paragraph{Datasets}
We test mixture models and baselines on three benchmark datasets that uniquely provide multiple human references~\citep{ott_icml18,hassan2018achieving}.

\textbf{WMT'17 English-German (En-De):}
We train on all available bitext
and filter sentence pairs that have source or target longer than 80 words, resulting in 4.5M sentence pairs.
We use the Moses tokenizer~\citep{koehn:moses:2007} and learn a joint source and target Byte-Pair-Encoding~\citep{sennrich:bpe:2016} with 32K types.
We develop on newstest2013 and test on a 500 sentence subset of newstest2014 that has 10 reference translations~\citep{ott_icml18}.

\textbf{WMT'14 English-French (En-Fr):}
We borrow the setup of~\citet{gehring2017convs2s} with 36M training sentence pairs and 40K joint BPE vocabulary. We validate on newstest2012+2013, and test on a 500 sentence subset of newstest2014 with 10 reference translations~\citep{ott_icml18}.

\textbf{WMT'17 Chinese-English (Zh-En):}
We pre-process the training data following \citet{hassan2018achieving} which results in 20M sentence pairs, 48K and 32K source and target BPE vocabularies respectively. We develop on devtest2017 and report results on newstest2017 with 3 reference translations.

\paragraph{Model Architecture}
All models use a very similar architecture, built using the  Transformer~\citep{vaswani2017attention} implementation in the Fairseq toolkit~\citep{ott2019fairseq}.
The encoder and decoder have 6 blocks.
The number of attention heads, embedding dimension and inner-layer dimension are 8, 512, 2048 for the ``base" configuration and 16, 1024, 4096 for the ``big" configuration, respectively.
We use the ``base" configuration to compare mixture model variants in \textsection\ref{subsec:res_moe}, and the ``big" configuration for extended experiments in \textsection\ref{subsec:eval_three_big}.

Mixture models with ``independent" experts use independent decoders, while models with ``shared" experts use a single shared decoder with an extra set of weights to embed each latent variable state.
All mixture models use a shared encoder across the experts.
Models that learn a prior, namely \smlp{} and \hmlp{}, have an additional module that averages the top encoder layer hidden states and predicts the conditional prior distribution $p(z|x;\theta)$ via a one hidden layer neural network with a tanh activation in between; the number of hidden units matches the embedding dimension.

\paragraph{Baselines}
The first baseline we consider use the same architecture, without a latent variable, and use beam search or sampling to generate $K$ hypotheses.
We also consider the following modified versions: \textsl{Diverse Beam Search}~\cite{vijayakumar2018diverse} and \textsl{Biased Sampling}~\cite{graves13,fan2018hierarchical,edunov2018bt}. 
The former performs beam search sequentially and penalizes the selection of words used in previous generations.
The latter improves upon straight sampling by sampling over the top-$k$ most likely words instead of all words at each step, resulting in less noisy generations.
Finally, we also compare against \textsl{Variational NMT}~\citep{zhang2016variational} which uses a Gaussian latent variable and variational inference for posterior approximation. We use a 512-dimensional latent space. At test time we first sample $z$ from the conditional prior distribution and then do greedy decoding.
Any additional hyper-parameters for these baselines are tuned via grid search over the validation set.

\paragraph{Experimental Details}
Models are optimized with the Adam algorithm \citep{kingma:adam:2015} using $\beta_1 = 0.9$, $\beta_2 = 0.98$, and $\epsilon = 1e-8$. We use the same learning rate schedule as \citet{ott:scaling:2018}.
We run experiments on between 8 and 128 Nvidia V100 GPUs with mini-batches of approximately 25K and 400K tokens for the experiments of \textsection\ref{subsec:res_moe} and \textsection\ref{subsec:eval_three_big}, respectively, following~\citet{ott:scaling:2018}.

\subsection{Analysis of Mixture Models: Tricks of the Trade}\label{subsec:moe_tricks}
In this section we compare the four variants of mixture models (\textsection\ref{subsec:model_variants}), and consider
for each such variant whether to share parameters among mixture components (\textsection\ref{subsec:parameterization}), whether to update responsibilities once each epoch (offline mode) or after every gradient step (online mode) (\textsection\ref{subsec:schedule}), and the effect of regularization on them (\textsection\ref{subsec:regularization}).
We train on the WMT'17 En-De dataset using the ``base" Transformer configuration and $K=3$ mixture components for efficiency considerations.

\begin{table}
\begin{center}
\small
\begin{tabular}{l|ccc|ccc}
\toprule
 & \multicolumn{3}{c|}{Pairwise-BLEU} & \multicolumn{3}{c}{BLEU} \\
 & no & E\&M & M & no & E\&M & M\\
\midrule
sMlp & \round{53.9} & \round{98.46} &  \round{57.8} & \round{58.24} & \round{66.40} & \round{64.4}   \\
sMup & \round{55.38} & \round{98.46} & \round{61.6} & \round{58.16} & \round{67.74} & \round{64.2} \\
hMlp & \round{45.88} & \round{89.31} & \round{47.8} & \round{54.65} & \round{65.85} & \round{60.3}  \\
hMup & \round{40.97} & \round{87.56} & \round{53.1} & \round{53.05} & \round{66.13} & \round{62.6} \\
\bottomrule
\end{tabular}
\end{center}
\caption{\small Results on the WMT'17 En-De dataset with $K=3$ for mixture models with no dropout (no), dropout$\,{=}\,0.1$ in both the E-step and M-step (E\&M), or in the M-step only (M).
For all model variants, dropout in the E-step causes the model to ignore the latent variable.
} 
\label{tab:regularization}
\end{table}

\paragraph{Effect of Dropout:}
Na\"ively optimizing the loss functions in Equation~\ref{eq:loss} with dropout noise causes mixture models to ignore the latent variable (D2).
This can be seen by the very high Pairwise-BLEU (low diversity) when dropout is used in both the E-step and M-step (Table~\ref{tab:regularization}, \texttt{E\&M}).
Unfortunately, entirely disabling dropout exacerbates overfitting and gives quite worse translation quality, i.e., lower BLEU (Table~\ref{tab:regularization}, \texttt{no}).
We provide a solution to this issue by decomposing the log-likelihood gradient as in Equation~\ref{eq:grads} and applying dropout only to the gradient computation of the experts (M-step) but not their responsibilities (E-step).
This allows the model to make consistent use of the experts and helps them to diversify (Table~\ref{tab:regularization}, \texttt{M}; see also Appendix~\ref{app:dropout}).
We adopt this dropout scheme for subsequent experiments.


\label{subsec:res_moe}
\begin{figure}[t]
\begin{center}
\includegraphics[width=0.99\columnwidth]{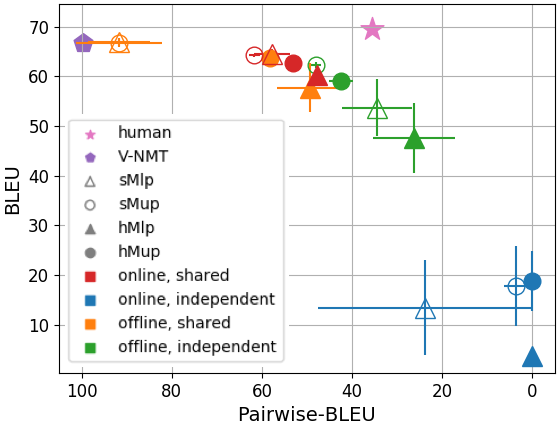}
\end{center}
\caption{\small Comparison of mixture models with different design choices on the WMT'17 En-De dataset with $K=3$ mixture components
(best viewed in color). Filled or empty markers show hard (h) or soft (s) mixture (M), and circle/triangle represent uniform prior (up)/learned prior (lp), respectively. Online/offline responsibility assignment and shared/independent parameterization are indicated by different colors.
The star marker represents human performance, and pentagon represents Variational NMT. Error bars are computed over five random training seeds.
Degeneracy D1 occurs in the lower right corner where some mixture components are hardly trained and produce poor generations. Degeneracy D2 is in the upper left, where the latent variable is ignored and different latent states produce almost the same hypotheses.
We strive for configurations close to human performance, which are of high quality as well as diverse (upper right).
}
\label{fig:moe_variants}
\end{figure}

\begin{figure*}[t]
\centering
\includegraphics[width=0.95\textwidth]{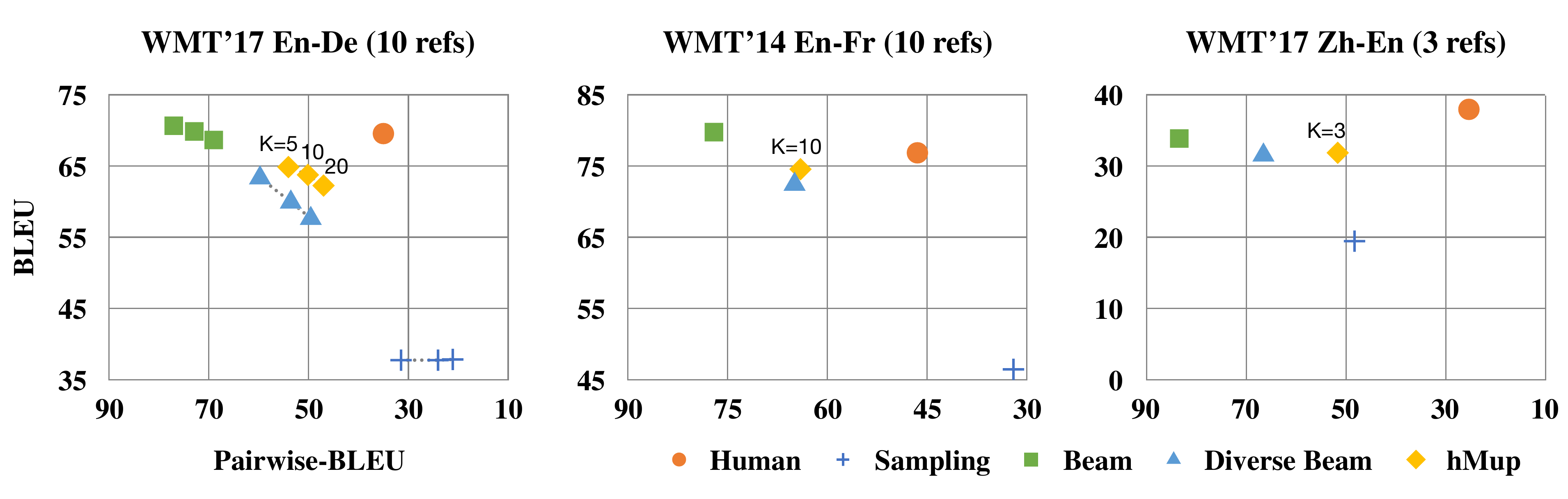}
\caption{\small Comparison of the \hmup{} mixture model with baselines on three WMT datasets. The mixture model provides the best trade-off between translation quality and diversity.
}
 \label{fig:plots}
\end{figure*}
\paragraph{Mixture Model Exploration:}
Figure~\ref{fig:moe_variants} shows the results of different modeling choices we introduced in \textsection\ref{sec:models}.
We plot Pairwise-BLEU in descending order versus BLEU, so the right side indicates high diversity and the upper side indicates high quality.
First, both Variational NMT (V-NMT) as well as soft mixture models with shared parameterization and offline responsibility updates lead to degeneracy D2 (upper left corner), where the latent variable is ignored and the generated hypotheses are almost the same, as indicated by Pairwise-BLEU close to 100.
This is a well-known failure mode of VAEs for text~\cite{bowman15}.\footnote{We also tried annealing the KL term for Variational NMT, but even then the latent variable is hardly used.}
In the mixture model case, we surmise that training a shared mixture for a long time with the initial (random) responsibilities prevents the latent variable embeddings from specializing, and thus they provide no useful information to the decoder.

Second, mixture models trained with independently parameterized components and online responsibility assignment are prone to degeneracy D1 (lower right corner), whereby only a single expert gets trained due to the ``rich gets richer" effect~\cite{largemoe17}, and BLEU dramatically drops because of poor generations from other experts.\footnote{We observe that the BLEU score is high for a single latent state but nearly zero for others, see Table~\ref{tab:giant_table} in Appendix~\ref{app:tables}.}
Parameter sharing alleviates this, as updates to one expert affect also the others.

Third, there are two settings that work consistently well: online responsibility update with shared parameters, and offline update with independent parameters.
The latter yields more diverse but lower quality translations, as expected since components have more degrees of freedom to deviate from each other but also less data to train.
Fourth and not surprisingly, soft assignment yields lower diversity than hard assignment~\cite{kearns1998information}.
Fifth, 
setting the prior to be uniform encourages the model to make use of all the components for each input source sentence, and thus gives better BLEU.

Overall, there are a few variants that work robustly as indicated by the small variance from random initialization, and strike a good balance between generation quality and diversity as indicated by proximity to human performance: all online-shared models, offline-shared \hmup{}, as well as offline-independent \smup{} and \hmup{}.
All of these models perform well, offering slightly different trade-offs between quality and diversity. 

If we also account for computational and memory cost, methods using shared parameters and hard EM are preferable, since the extra parameters are negligible and it requires only a single backward pass for the selected expert.\footnote{With $K=10$ and the ``base'' Transformer architecture, hard mixture models are 2.4 times faster than the soft counterpart.}
If we further consider simplicity of implementation, uniform prior and online responsibility update are favorable because they do not require additional model components, nor storing the responsibilities as for offline responsibility update.
Taking these considerations into account, we adopt \hmup{} (online-shared) for subsequent experiments, keeping in mind that the model variants mentioned above are likely to work similarly well.

\subsection{Large-Scale Evaluation} \label{subsec:eval_three_big}

\begin{table}[t]
\def\arraystretch{1.1}
\small
\begin{center}
\begin{tabular}{l|ccc|ccc}
\toprule
 & \multicolumn{3}{c|}{Pairwise-BLEU} & \multicolumn{3}{c}{BLEU} \\
 & en-de & en-fr & zh-en & en-de & en-fr & zh-en \\
\midrule
sampling & \round{24.14} & \round{32.02} & \round{48.23} & \round{37.8} & \round{46.49} & \round{19.47} \\
beam & \round{73.02} & \round{77.06} & \round{83.35} & \round{69.93} & \round{79.8} & \round{33.94} \\
div-beam & \round{53.66} & \round{64.93} & \round{66.48} & \round{59.96} & \round{72.54} & \round{31.6} \\
\hmup{} & \round{50.24} & \round{64.04} & \round{51.59} & \round{63.76} & \round{74.58} & \round{31.92} \\
\hline
human & \round{35.48} & \round{46.46} & \round{25.29} & \round{69.63} & \round{76.87} & \round{38.03} \\
\toprule
\end{tabular}
\end{center}
\caption{\small Results on three WMT datasets: En-De, En-Fr, Zh-En have 10, 10, 3 human references respectively.
We generate the same number of hypotheses as the number of references.
}\label{tab:results}
\end{table}

We now extend our evaluation to three large benchmark datasets using the ``big" Transformer configuration and a larger numbers of mixture components ($K=10,10,3$ for WMT En-De, En-Fr and Zh-En datasets respectively, to match the number of references we have on the test set).

Table~\ref{tab:results} and Figure~\ref{fig:plots} show results of the \hmup{} mixture model compared to other baseline approaches.
The BLEU of beam search is fairly close to human, implying a remarkably high generation quality.
However, it severely lacks diversity, as Pairwise-BLEU is close to 80.
This suggests a scenario similar to Figure~\ref{fig:toy_metrics}-B.
In contrast, sampling hypotheses are very diverse, typically even more so than human references on En-De and En-Fr, but have a very poor BLEU, as illustrated in Figure~\ref{fig:toy_metrics}-A.
Diverse beam search increases the diversity of beam search to some extent, but pays a cost in translation quality.
Overall, \hmup{} achieves the best trade-off between quality and diversity.

We also explore the impact of varying $K$ on the WMT'17 En-De dataset (Figure~\ref{fig:plots}, \emph{left}).
In general, when more hypotheses are generated, they become more diverse but of worse quality.
Despite the improvement offered by \hmup{}, there is still a large gap between the diversity of human references and system generations.
The complete results of these experiments and the biased sampling baseline are provided in Appendix~\ref{app:tables}.

\subsection{Qualitative Analysis}

\begin{table*}[t]
    \def\arraystretch{1.1}\setlength{\tabcolsep}{3pt}
    \scriptsize
    \centering
    \begin{tabular}{lll}
           \toprule
           Source & \begin{CJK*}{UTF8}{gbsn}参与 投票 的 成员 中 ， 58\% 反对 该 合同 交易 。\end{CJK*} & \begin{CJK*}{UTF8}{gbsn}自 11 月份 开始 ， 俄罗斯 民意 也 有所 扭转 。\end{CJK*}\\
           References & It was rejected by 58 \% of its members who voted in the ballot . & Russian public opinion has also turned since November .\\
           & Of the members who voted , 58 \% opposed the contract transaction . & Russian public opinion has started to change since November .\\
           & Of the members who participated in the vote , 58 \% opposed the contract . & The polls in Russian show a twist turn since the beginning of November .\\
           \\
           Beam & Fifty-eight per cent of those voting opposed the contract deal . & Since November , public opinion in Russia has also shifted .\\
           & Fifty-eight per cent of the voting members opposed the contract deal . & Since November , public opinion in Russia has also reversed .\\
           & Fifty-eight per cent of the voting members opposed the contract transaction . & Since November , Russian public opinion has also shifted .\\
           \\
           Diverse beam & Of the members voting , 58 per cent opposed the contract deal . & Since November , public opinion in Russia has also shifted .\\
            & Of the members voting , 58 per cent opposed the contract deal . & Since November , the mood in Russia has also changed .\\
           & Of the members voting , 58 per cent opposed the transaction . & Since November , public opinion in Russia has also been reversed .\\
           \\
           \hmup{} & Fifty-eight per cent of the members who voted opposed the contract deal . & Since November , opinion in Russia has also reversed .\\
            & Of the members who voted , 58 \% opposed the deal . & Since November , the mood in Russia has also reversed .\\
           & Fifty-eight per cent of the voting members opposed the contract deal . & Opinion in Russia has also shifted since November .\\
           \toprule
    \end{tabular}
    \caption{\small Examples of generations by the \hmup{} mixture model and baselines on the WMT'17 Zh-En dataset. The mixture model shows considerable diversity compared to beam search and diverse beam search.}
    \label{tab:examples_all}
\end{table*}
\begin{table*}[t]
    \def\arraystretch{1.1}
    \scriptsize
    \centering
    \begin{tabular}{lll}
         \toprule
         Source & \begin{CJK*}{UTF8}{gbsn}这是 一次 规模 巨大 的 作业 ， 同时 也 是 一次 非常 精密 的 作业 。\end{CJK*} & \begin{CJK*}{UTF8}{gbsn}他 从不 愿意 与 家人 争吵 。\end{CJK*}\\
         Reference & This was a massive and , at the same time , very delicate operation . & He never wanted to be in any kind of altercation .\\
         hMup & It was a huge job , and a very delicate one as well . & He never liked to quarrel with his family .\\
          & It 's a very large job , and it 's a very delicate one , too . & He never wants to quarrel with his family\\
         & This is a huge job , but also a very delicate one . & He never likes to argue with his family\\
         \\
         Source & \begin{CJK*}{UTF8}{gbsn}不断 的 恐怖袭击 显然 已 对 他 造成 很大 打击 。\end{CJK*} & \begin{CJK*}{UTF8}{gbsn}我 不想 说 这 是 我 的 最后 一场 比赛 。\end{CJK*}\\
         Reference & Repeat terror attacks on Turkey have clearly shaken him too . & I didn 't want to say this was my last race .\\
         hMup & The continuing terrorist attacks had apparently hit him hard . & I didn 't want to say it was my last game .\\
          & He is clearly already being hit hard by the continuing terrorist attacks .  & I don 't want to say it 's my last game , he said .\\
         & Repeated terrorist attacks have apparently hit him hard . & I don 't want to say this is my last game .\\
         \\
         Source & \begin{CJK*}{UTF8}{gbsn}由此 判断 ， 这 无疑 是 一场 持续 战 。\end{CJK*} & \begin{CJK*}{UTF8}{gbsn}两人 2015 年 缴纳 了 20.3\% 的 联邦 税 。\end{CJK*}\\
         Reference & It appears that this was definitely an ongoing battle . & They paid a federal effective tax rate of 20.3 percent in 2015 .\\
         hMup & Judging by that , it is undoubtedly a continuing battle . & Both paid a federal tax of 20.3 per cent in 2015 .\\
          & It is a battle that is no doubt ongoing . & They paid a federal tax of 20.3 \% in 2015 .\\
         & Judging by this , this is undoubtedly a continuing battle . & Both paid 20.3 \% of federal taxes in 2015 .\\

         \toprule
    \end{tabular}
    \caption{\small More examples of the \hmup{} mixture model on the WMT'17 Zh-En dataset. Different latent values learn to specialize for different translation styles consistently across examples, such as past tense vs. present tense, \texttt{this} vs. \texttt{that}, and \texttt{per cent} vs. \texttt{\%}. }
    \label{tab:examples_hMoE}
\end{table*}

In this section we perform a qualitative analysis with the WMT'17 Zh-En dataset.
In Table~\ref{tab:examples_all} we show two source sentences, the corresponding reference translations, and generated hypotheses from different approaches.
We see that beam search tends to produce generations that differ only in the last few words.
Diverse beam search improves the diversity over beam search, but is not as diverse as \hmup{} and may produce duplicate hypotheses (e.g., if the diversity penalty is not sufficiently high).
\hmup{} shows significant diversity in wording, word order, clause structure, etc.

To investigate whether the latent variable in \hmup{} learns different translation styles, we examine the hypotheses generated from each latent state.
For each value of $z$, we compute word frequencies of the corresponding generations and look for words whose frequency is significantly different as we change the value of the latent variable. We first discover that for words like \texttt{was}, \texttt{were} and \texttt{had}, $z\,{=}\,1$'s frequency is more than three times higher than $z\,{=}\,3$'s; conversely, for \texttt{has} and \texttt{says}, $z\,{=}\,3$'s frequency is more than twice higher than $z\,{=}\,1$'s. 
Since Chinese does not have tense unless a time phrase is explicitly stated, we speculate that when translating into English, the first latent value tends to translate with past tense whereas the third latent value tends to translate with present tense. Indeed we find that this is a consistent behavior, as seen from the first four examples in Table~\ref{tab:examples_hMoE}.
Similarly, we find that different latent values exhibit different preferences for using \texttt{this} or \texttt{that} (see the fourth and fifth examples in  Table~\ref{tab:examples_hMoE}), \texttt{\%} or \texttt{per cent} (see the last example in Table~\ref{tab:examples_hMoE} and the first example in Table~\ref{tab:examples_all}), and so on. 

\section{Conclusion} \label{sec:conclusion}
Using large scale benchmarks and state-of-the-art architectures, we investigated $32$ variants of mixture models, arising from the following design choices: use of hard versus soft EM, uniform versus learned prior, shared versus independent parameterization of mixture components, online versus offline responsibility update, and use of standard dropout regularization versus removal of this noise in responsibility computation.
To the best of our knowledge this is the most extensive study of mixture models for conditional text generation to date, with machine translation as a use case.
The simplicity of mixture models provides important advantages: the latent variable assignment can be explicitly enumerated and the posterior can be computed exactly.
Despite their simplicity, however, mixture models exhibit complex behaviors depending on different combinations of design choices. In particular, when instantiated with sub-optimal choices, they are prone to two typical failure modes---only one component gets trained and other components ``die'', or the latent variable is ignored and all components behave the same.
Our study provides insights into training deep sequence models with discrete latent variables. 

Our recommended configurations enable mixture models to offer much better trade-offs between quality and diversity than variational models as well as heuristic diverse decoding approaches.
In the future, we hope to broaden the scope of this work by looking at other generation tasks such as dialogue and image captioning.
We would also like to investigate models with richer and more structured latent representations, and narrow the gap between model and human performance.

\subsection*{Acknowledgements}
We thank David Grangier for insightful discussions in the preliminary phase of this work. We also thank MIT NLP group for their helpful comments.
\bibliography{main}

\begin{thebibliography}{37}
\providecommand{\natexlab}[1]{#1}
\providecommand{\url}[1]{\texttt{#1}}
\expandafter\ifx\csname urlstyle\endcsname\relax
  \providecommand{\doi}[1]{doi: #1}\else
  \providecommand{\doi}{doi: \begingroup \urlstyle{rm}\Url}\fi

\bibitem[Bowman et~al.(2016)Bowman, Vilnis, Vinyals, Dai, Jozefowicz, and
  Bengio]{bowman15}
Bowman, S.~R., Vilnis, L., Vinyals, O., Dai, A.~M., Jozefowicz, R., and Bengio,
  S.
\newblock Generating sentences from a continuous space.
\newblock In \emph{SIGNLL Conference on Computational Natural Language
  Learning}, 2016.

\bibitem[Cao \& Clark(2017)Cao and Clark]{cao2017latent}
Cao, K. and Clark, S.
\newblock Latent variable dialogue models and their diversity.
\newblock In \emph{Proceedings of EACL (Short Papers)}, pp.\  182--187, 2017.

\bibitem[Dai et~al.(2017)Dai, Fidler, Urtasun, and Lin]{dai2017towards}
Dai, B., Fidler, S., Urtasun, R., and Lin, D.
\newblock Towards diverse and natural image descriptions via a conditional gan.
\newblock In \emph{2017 IEEE International Conference on Computer Vision
  (ICCV)}, pp.\  2989--2998. IEEE, 2017.

\bibitem[Dempster et~al.(1977)Dempster, Laird, and Rubin]{dempster1977maximum}
Dempster, A.~P., Laird, N.~M., and Rubin, D.~B.
\newblock Maximum likelihood from incomplete data via the em algorithm.
\newblock \emph{Journal of the royal statistical society. Series B
  (methodological)}, pp.\  1--38, 1977.

\bibitem[Dreyer \& Marcu(2012)Dreyer and Marcu]{hyter}
Dreyer, M. and Marcu, D.
\newblock Hyter: Meaning-equivalent semantics for translation evaluation.
\newblock In \emph{ACL}, 2012.

\bibitem[Edunov et~al.(2018)Edunov, Ott, Auli, and Grangier]{edunov2018bt}
Edunov, S., Ott, M., Auli, M., and Grangier, D.
\newblock Understanding back-translation at scale.
\newblock In \emph{Proc. of EMNLP}, 2018.

\bibitem[Eigen et~al.(2014)Eigen, Ranzato, and Sutskever]{eigen14}
Eigen, D., Ranzato, M., and Sutskever, I.
\newblock Learning factored representations in a deep mixture of experts.
\newblock \emph{ICLR}, 2014.

\bibitem[Fan et~al.(2018)Fan, Lewis, and Dauphin]{fan2018hierarchical}
Fan, A., Lewis, M., and Dauphin, Y.
\newblock Hierarchical neural story generation.
\newblock In \emph{Proceedings of the 56th Annual Meeting of the Association
  for Computational Linguistics (Long Papers)}, pp.\  889--898. Association for
  Computational Linguistics, 2018.

\bibitem[Galley et~al.(2015)Galley, Brockett, Sordoni, Ji, Auli, Quirk,
  Mitchell, Gao, and Dolan]{galley2015deltableu}
Galley, M., Brockett, C., Sordoni, A., Ji, Y., Auli, M., Quirk, C., Mitchell,
  M., Gao, J., and Dolan, B.
\newblock deltableu: A discriminative metric for generation tasks with
  intrinsically diverse targets.
\newblock In \emph{Proceedings of ACL (Short Papers)}, pp.\  445--450, 2015.

\bibitem[Gehring et~al.(2017)Gehring, Auli, Grangier, Yarats, and
  Dauphin]{gehring2017convs2s}
Gehring, J., Auli, M., Grangier, D., Yarats, D., and Dauphin, Y.~N.
\newblock {Convolutional Sequence to Sequence Learning}.
\newblock In \emph{Proc. of ICML}, 2017.

\bibitem[Graves(2013)]{graves13}
Graves, A.
\newblock Generating sequences with recurrent neural networks.
\newblock \emph{arXiv:1308.0850}, 2013.

\bibitem[Guzman-Rivera et~al.(2012)Guzman-Rivera, Batra, and
  Kohli]{guzman2012multiple}
Guzman-Rivera, A., Batra, D., and Kohli, P.
\newblock Multiple choice learning: Learning to produce multiple structured
  outputs.
\newblock In \emph{Advances in Neural Information Processing Systems}, pp.\
  1799--1807, 2012.

\bibitem[Hassan et~al.(2018)Hassan, Aue, Chen, Chowdhary, Clark, Federmann,
  Huang, Junczys-Dowmunt, Lewis, Li, et~al.]{hassan2018achieving}
Hassan, H., Aue, A., Chen, C., Chowdhary, V., Clark, J., Federmann, C., Huang,
  X., Junczys-Dowmunt, M., Lewis, W., Li, M., et~al.
\newblock Achieving human parity on automatic chinese to english news
  translation.
\newblock \emph{arXiv preprint arXiv:1803.05567}, 2018.

\bibitem[He et~al.(2018)He, Haffari, and Norouzi]{he2018sequence}
He, X., Haffari, G., and Norouzi, M.
\newblock Sequence to sequence mixture model for diverse machine translation.
\newblock In \emph{Proceedings of the 22nd Conference on Computational Natural
  Language Learning}, pp.\  583--592, 2018.

\bibitem[Jacobs et~al.(1991)Jacobs, Jordan, Nowlan, , and Hinton]{moe91}
Jacobs, R.~A., Jordan, M.~I., Nowlan, S., , and Hinton, G.~E.
\newblock Adaptive mixtures of local experts.
\newblock \emph{Neural Computation}, 3:\penalty0 1–--12, 1991.

\bibitem[Kaiser et~al.(2018)Kaiser, Bengio, Roy, Vaswani, Parmar, Uszkoreit,
  and Shazeer]{kaiser18}
Kaiser, L., Bengio, S., Roy, A., Vaswani, A., Parmar, N., Uszkoreit, J., and
  Shazeer, N.
\newblock Fast decoding in sequence models using discrete latent variables.
\newblock In \emph{International Conference on Machine Learning}, pp.\
  2395--2404, 2018.

\bibitem[Kearns et~al.(1998)Kearns, Mansour, and Ng]{kearns1998information}
Kearns, M., Mansour, Y., and Ng, A.~Y.
\newblock An information-theoretic analysis of hard and soft assignment methods
  for clustering.
\newblock In \emph{Learning in graphical models}, pp.\  495--520. Springer,
  1998.

\bibitem[Kingma \& Ba(2015)Kingma and Ba]{kingma:adam:2015}
Kingma, D.~P. and Ba, J.
\newblock {Adam: A Method for Stochastic Optimization}.
\newblock In \emph{International Conference on Learning Representations
  ({ICLR})}, 2015.

\bibitem[Kingma \& Welling(2014)Kingma and Welling]{kingma2014auto}
Kingma, D.~P. and Welling, M.
\newblock Auto-encoding variational bayes.
\newblock \emph{Proceedings of the International Conference on Learning
  Representations (ICLR)}, 2014.

\bibitem[Koehn et~al.(2007)Koehn, Hoang, Birch, Callison-Burch, Federico,
  Bertoldi, Cowan, Shen, Moran, Zens, Dyer, Bojar, Constantin, and
  Herbst]{koehn:moses:2007}
Koehn, P., Hoang, H., Birch, A., Callison-Burch, C., Federico, M., Bertoldi,
  N., Cowan, B., Shen, W., Moran, C., Zens, R., Dyer, C., Bojar, O.,
  Constantin, A., and Herbst, E.
\newblock Moses: Open source toolkit for statistical machine translation.
\newblock In \emph{ACL Demo Session}, 2007.

\bibitem[Lee et~al.(2016)Lee, Purushwalkam, Cogswell, Ranjan, Crandall, and
  Batra]{lee16}
Lee, S., Purushwalkam, S., Cogswell, M., Ranjan, V., Crandall, D., and Batra,
  D.
\newblock Stochastic multiple choice learning for training diverse deep
  ensembles.
\newblock In \emph{NIPS}, 2016.

\bibitem[Li et~al.(2017)Li, Monroe, and Jurafsky]{jiweili17}
Li, J., Monroe, W., and Jurafsky, D.
\newblock A simple, fast diverse decoding algorithm for neural generation.
\newblock \emph{arXiv:1611.08562v2}, 2017.

\bibitem[Neal \& Hinton(1998)Neal and Hinton]{neal1998view}
Neal, R.~M. and Hinton, G.~E.
\newblock A view of the em algorithm that justifies incremental, sparse, and
  other variants.
\newblock In \emph{Learning in graphical models}, pp.\  355--368. Springer,
  1998.

\bibitem[Ott et~al.(2018{\natexlab{a}})Ott, Auli, Grangier, and
  Ranzato]{ott_icml18}
Ott, M., Auli, M., Grangier, D., and Ranzato, M.
\newblock Analyzing uncertainty in neural machine translation.
\newblock In \emph{International Conference of Machine Learning},
  2018{\natexlab{a}}.

\bibitem[Ott et~al.(2018{\natexlab{b}})Ott, Edunov, Grangier, and
  Auli]{ott:scaling:2018}
Ott, M., Edunov, S., Grangier, D., and Auli, M.
\newblock Scaling neural machine translation.
\newblock In \emph{Proceedings of the Third Conference on Machine Translation:
  Research Papers}, 2018{\natexlab{b}}.

\bibitem[Ott et~al.(2019)Ott, Edunov, Baevski, Fan, Gross, Ng, Grangier, and
  Auli]{ott2019fairseq}
Ott, M., Edunov, S., Baevski, A., Fan, A., Gross, S., Ng, N., Grangier, D., and
  Auli, M.
\newblock fairseq: A fast, extensible toolkit for sequence modeling.
\newblock In \emph{Proceedings of NAACL-HLT 2019: Demonstrations}, 2019.

\bibitem[Papineni et~al.(2002)Papineni, Roukos, Ward, and
  Zhu]{papineni2002bleu}
Papineni, K., Roukos, S., Ward, T., and Zhu, W.-J.
\newblock Bleu: a method for automatic evaluation of machine translation.
\newblock In \emph{Proceedings of the 40th annual meeting on association for
  computational linguistics}, pp.\  311--318. Association for Computational
  Linguistics, 2002.

\bibitem[Schulz et~al.(2018)Schulz, Aziz, and Cohn]{schulz2018stochastic}
Schulz, P., Aziz, W., and Cohn, T.
\newblock A stochastic decoder for neural machine translation.
\newblock \emph{Proceedings of the 56th Annual Meeting of the Association for
  Computational Linguistics}, 2018.

\bibitem[Sennrich et~al.(2016)Sennrich, Haddow, and Birch]{sennrich:bpe:2016}
Sennrich, R., Haddow, B., and Birch, A.
\newblock Neural machine translation of rare words with subword units.
\newblock In \emph{Conference of the Association for Computational Linguistics
  {(ACL)}}, 2016.

\bibitem[Serban et~al.(2017)Serban, Sordoni, Lowe, Charlin, Pineau, Courville,
  and Bengio]{serban2017hierarchical}
Serban, I.~V., Sordoni, A., Lowe, R., Charlin, L., Pineau, J., Courville,
  A.~C., and Bengio, Y.
\newblock A hierarchical latent variable encoder-decoder model for generating
  dialogues.
\newblock In \emph{AAAI}, pp.\  3295--3301, 2017.

\bibitem[Shazeer et~al.(2017)Shazeer, Mirhoseini, Maziarz, Davis, Le, Hinton,
  and Dean]{largemoe17}
Shazeer, N., Mirhoseini, A., Maziarz, K., Davis, A., Le, Q., Hinton, G., and
  Dean, J.
\newblock Outrageously large neural networks: The sparsely-gated
  mixture-of-experts layer.
\newblock In \emph{International Conference on Learning Representations}, 2017.

\bibitem[Vaswani et~al.(2017)Vaswani, Shazeer, Parmar, Uszkoreit, Jones, Gomez,
  Kaiser, and Polosukhin]{vaswani2017attention}
Vaswani, A., Shazeer, N., Parmar, N., Uszkoreit, J., Jones, L., Gomez, A.~N.,
  Kaiser, {\L}., and Polosukhin, I.
\newblock Attention is all you need.
\newblock In \emph{Advances in Neural Information Processing Systems}, pp.\
  5998--6008, 2017.

\bibitem[Vijayakumar et~al.(2018)Vijayakumar, Cogswell, Selvaraju, Sun, Lee,
  Crandall, and Batra]{vijayakumar2018diverse}
Vijayakumar, A.~K., Cogswell, M., Selvaraju, R.~R., Sun, Q., Lee, S., Crandall,
  D., and Batra, D.
\newblock Diverse beam search for improved description of complex scenes.
\newblock In \emph{Thirty-Second AAAI Conference on Artificial Intelligence},
  2018.

\bibitem[Wang et~al.(2017)Wang, Schwing, and Lazebnik]{wang2017diverse}
Wang, L., Schwing, A., and Lazebnik, S.
\newblock Diverse and accurate image description using a variational
  auto-encoder with an additive gaussian encoding space.
\newblock In \emph{Advances in Neural Information Processing Systems}, pp.\
  5756--5766, 2017.

\bibitem[Wen et~al.(2017)Wen, Miao, Blunsom, and Young]{wen2017latent}
Wen, T.-H., Miao, Y., Blunsom, P., and Young, S.
\newblock Latent intention dialogue models.
\newblock In \emph{Proceedings of the 34th International Conference on Machine
  Learning}, pp.\  3732--3741, 2017.

\bibitem[Yu et~al.(2017)Yu, Zhang, Wang, and Yu]{yu2017seqgan}
Yu, L., Zhang, W., Wang, J., and Yu, Y.
\newblock Seqgan: Sequence generative adversarial nets with policy gradient.
\newblock In \emph{Thirty-First AAAI Conference on Artificial Intelligence},
  2017.

\bibitem[Zhang et~al.(2016)Zhang, Xiong, Su, Duan, and
  Zhang]{zhang2016variational}
Zhang, B., Xiong, D., Su, J., Duan, H., and Zhang, M.
\newblock Variational neural machine translation.
\newblock \emph{Proceedings of the 2016 Conference on Empirical Methods in
  Natural Language Processing}, 2016.

\end{thebibliography}
\bibliographystyle{icml2019}

\newpage
\clearpage
\appendix

\section{Effect of Dropout}\label{app:dropout}

In Section~\ref{subsec:moe_tricks} we observed that it is crucial to turn off dropout during the computation of responsibilities (E-step) to avoid model collapse, see Table~\ref{tab:regularization}.
In this section, we further investigate the impact of dropout on the responsibility computation, using \hmup{} as an example.

We speculate that dropout noise weakens the dependency on the latent variable, causing the hard E-step to select among the latent values at random.
This prevents different latent states from specializing and ultimately causes the model to ignore them.

\begin{figure}[h]
    \centering
    \includegraphics[width=0.85\columnwidth]{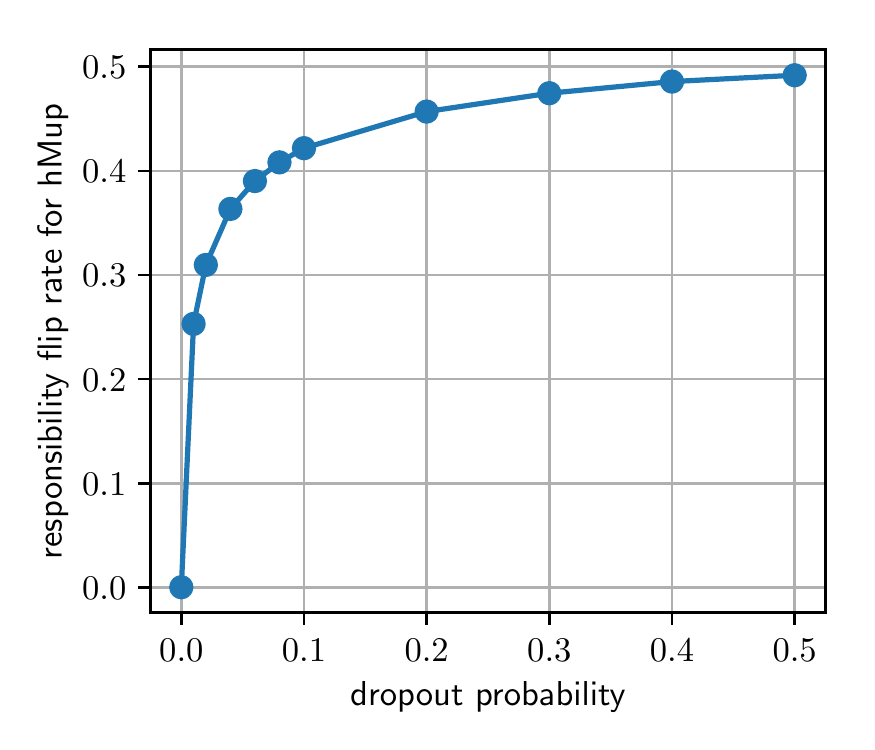}
    \caption{\small The effect of dropout on the E-step at the beginning of training. On the x-axis the dropout rate, and on the y-axis the fraction of times that the (hard) responsibility assignment is changed when dropout noise is turned on. Even a small amount of dropout noise causes excessive randomness of responsibility assignment, which in turn makes the model  ignore the latent variable. It is therefore important to turn off dropout when estimating responsibilities. 
    Experiments are performed on the WMT'17 En-De dataset with 2 latent categories ($K=2$) and the ``base" Transformer architecture.}
     \label{fig:effect_of_dropout}
\end{figure}

To test this hypothesis, we show in Figure~\ref{fig:effect_of_dropout} the effect of dropout noise on the E-step at the beginning of training (i.e., with a randomly initialized model).
On the y-axis we plot how often the optimal value of $z$ changes after applying dropout with different rates.
We see that as we increase the dropout probability, the optimal value of $z$ is quickly corrupted---even with a small dropout probability of 0.1 we observe a 42\% chance that the optimal assignment of $z$ changes.

\section{Another Diversity Metric}\label{app:metric}
In addition to Pairwise-BLEU, we also consider another metric to evaluate the diversity of a set of hypotheses, the  \textbf{reference coverage}. 

We pair each hypothesis to its best matching reference (breaking ties randomly), count how many distinct references are matched to at least one hypothesis, and average this count over all sentences in the test set. A low coverage number indicates that all hypotheses are close to a few references. Instead, we would like a diverse set that covers most of the references.

Compared to Pairwise-BLEU, this metric offers more intuitive numerical values, as it ranges between 1 and the total number of available references. In the next section, we will report both values for completeness. 

\section{Detailed Results}\label{app:tables}

Table~\ref{tab:fullresults} compares two well performing mixture model variants, namely (online, shared) \hmup{} and \smup{}, to several baselines on the three banchmark datasets we have considered in this work, expanding on the results reported in  Table~\ref{tab:results} and Figure~\ref{fig:plots} of the main paper. Once again, mixture models offer a good trade-off between translation quality and diversity.

In Table~\ref{tab:varyk} we compare different approaches for generating diverse translations on the WMT'17 En-De dataset.
We additionally compare each approach as we vary the number of desired translations ($K$) (see also Figure~\ref{fig:plots}, \emph{left}).
We observe that sampling produces diverse but low quality outputs.
We can improve translation quality by restricting sampling to the top-$k$ candidates at each output position ($k\,{=}\,2$ performed best), but translation quality is still worse than \hmup{}.
Beam search produces the highest quality outputs, but with low diversity.
Diverse beam search provides a reasonable balance between diversity and translation quality, but \hmup{} produces even more diverse and better quality translations.
Finally, except for unrestricted sampling, \hmup{} covers the largest number of references among all the approaches evaluated. 

We conclude with  Table~\ref{tab:giant_table} which shows the values used to generate Figure~\ref{fig:moe_variants},  together with other metrics, such as corpus level BLEU with hypotheses generated by a fixed latent variable state throughout the whole test set. This metric is  useful to detect models affected by degeneracy D1, as there will be states that yield very low corpus level BLEU because they rarely generate good hypotheses.

\begin{table*}[t]
\def\arraystretch{1.1}
\small
\begin{center}
\begin{tabular}{l|ccc|ccc|ccc}
\toprule
 & \multicolumn{3}{c|}{Pairwise-BLEU} & \multicolumn{3}{c|}{BLEU} & \multicolumn{3}{c}{\#refs covered}\\
 & en-de & en-fr & zh-en & en-de & en-fr & zh-en & en-de & en-fr & zh-en\\
\midrule
Sampling & \round{24.14} & \round{32.02} & \round{48.23} & \round{37.8} & \round{46.49} & \round{19.47} & \round{4.63} & \round{4.33} & \round{1.49} \\
Beam & \round{73.02} & \round{77.06} & \round{83.35} & \round{69.93} & \round{79.8} & \round{33.94} & \round{3.13} & \round{3.17} & \round{1.27} \\
Diverse beam & \round{53.66} & \round{64.93} & \round{66.48} & \round{59.96} & \round{72.54} & \round{31.6} & \round{3.66} & \round{3.46} & \round{1.38} \\
sMup~\citep{he2018sequence} & \round{68.93} & \round{80.36} & \round{60.88} & \round{68.1} & \round{79.58} & \round{32.8} & \round{2.89} & \round{2.68} & \round{1.43} \\
hMup & \round{50.24} & \round{64.04} & \round{51.59} & \round{63.76} & \round{74.58} & \round{31.92} & \round{3.99} & \round{3.66} & \round{1.55} \\
\hline
Human & \round{35.48} & \round{46.46} & \round{25.29} & \round{69.63} & \round{76.87} & \round{38.03} & - & - & -\\
\toprule
\end{tabular}
\end{center}
\caption{\small Results on three WMT datasets. Extended version of Table~\ref{tab:results}, including the results of another mixture model \smup{}~\citep{he2018sequence}
and the reference coverage metric. \hmup{} and \smup{} provide different trade-offs between quality and diversity: the former is more diverse (lower Pairwise-BLEU), while the latter gives higher translation quality (BLEU).}\label{tab:fullresults}
\end{table*}

\begin{table*}[t]
\def\arraystretch{1.1}
\small
\begin{center}
\begin{tabular}{l|ccc|ccc|ccc}
\toprule
& \multicolumn{3}{c|}{Pairwise-BLEU} & \multicolumn{3}{c|}{BLEU} & \multicolumn{3}{c}{\#refs covered}\\
 & $K=5$ & 10 & 20 & 5 & 10 & 20 & 5 & 10 & 20\\
\midrule
Sampling & \round{31.55} & \round{24.14} & \round{21.19} & \round{37.83} & \round{37.8} & \round{37.94} & \round{3.05} & \round{4.63} & \round{6.23} \\
Biased sampling (top-2) & \round{49.25} & \round{47.84} & \round{46.70} & \round{59.69} & \round{59.98} & \round{60.35} & \round{2.71} & \round{3.65} & \round{4.72} \\
Beam & \round{77.13} & \round{73.02} & \round{69.11} & \round{70.73} & \round{69.93} & \round{68.74} & \round{2.31} & \round{3.13} & \round{3.99} \\
Diverse beam & \round{59.84} & \round{53.66} & \round{49.66} & \round{63.37} & \round{59.96} & \round{57.7} & \round{2.54} & \round{3.66} & \round{4.82}\\
hMup & \round{54.15} & \round{50.24} & \round{47.10} & \round{64.94} & \round{63.76} & \round{62.34} & \round{2.82} & \round{3.99} & \round{5.30}\\

\toprule
\end{tabular}
\end{center}
\caption{\small Results on the WMT'17 En-De dataset with various numbers of generations ($K$). We compare: multinomial sampling (\texttt{Sampling}); sampling restricted to the top-$k$ candidates at each step (\texttt{Biased sampling (top-2)}; $k\,{=}\,2$ performed best); beam search with varying beam widths (\texttt{Beam}); diverse beam search~\citep{vijayakumar2018diverse} with varying number of outputs (\texttt{Diverse beam}; note that the number of groups $G$ and diversity strength are tuned separately for each value of $K$); and the \hmup{} mixture model with $K$ components (\texttt{\hmup{}}).
} \label{tab:varyk}
\end{table*}

\begin{table*}[t]
\def\arraystretch{1.1}
\small
\begin{center}
\begin{tabular}{lll|cccccc}
\toprule
\multirow{2}{*}{schedule} & \multirow{2}{*}{parameterization} & \multirow{2}{*}{loss} & \multicolumn{3}{c}{BLEU per latent} & \multirow{2}{*}{Pairwise-BLEU} &	\multirow{2}{*}{BLEU} & \multirow{2}{*}{\#refs covered} \\
&&& $z=1$ & 2 & 3 \\
\midrule
online & shared & sMlp & 25.9 & 24.9 & 22.6 & 57.8 (4.0) & 64.4 (1.0) & 1.9 \\
online & shared & sMup & 25.8 & 25.2 & 22.8 & 61.6 (1.3) & 64.2 (0.4) & 1.9 \\
online & shared & hMlp & 25.5 & 22.6 & 21.5 & 47.8 (0.6) & 60.3 (0.3) & 2.1 \\
online & shared & hMup & 25.6 & 24.4 & 21.3 & 53.1 (1.2) & 62.6 (0.6) & 2.1 \\
\hline
online & independent & sMlp & 25.5 & {\color{red}0.0} & {\color{red}0.0} & 23.8 (23.8) & 13.4 (9.6) & 2.6 \\
online & independent & sMup & 25.8 & {\color{red}0.6} & {\color{red}0.0} & 3.5 (2.9) & 17.8 (8.1) & 2.6 \\
online & independent & hMlp & 26.1 & {\color{red}0.0} & {\color{red}0.0} & 0.0 (0.0) & 3.7 (0.0) & 2.7 \\
online & independent & hMup & 25.7 & {\color{red}0.2} & {\color{red}0.0} & 0.1 (0.1) & 18.7 (6.0) & 2.4 \\
\hline
offline & shared & sMlp & 25.8 & 25.7 & 25.6 & {\color{blue}91.7} (6.8) & 66.8 (0.7) & 1.4 \\
offline & shared & sMup & 25.7 & 25.5 & 25.3 & {\color{blue}91.7} (9.6) & 66.8 (0.9) & 1.4 \\
offline & shared & hMlp & 25.6 & 21.2 & 19.4 & 49.2 (7.3) & 57.7 (5.0) & 2.1 \\
offline & shared & hMup & 25.3 & 24.4 & 23.6 & 58.2 (1.5) & 63.7 (0.6) & 2.0 \\
\hline
offline & independent & sMlp & 25.7 & 19.4 & 16.4 & 34.4 (7.8) & 53.6 (5.7) & 2.2 \\
offline & independent & sMup & 25.3 & 23.5 & 22.9 & 48.0 (1.1) & 62.3 (0.6) & 2.1 \\
offline & independent & hMlp & 25.7 & 15.8 & 13.0 & 26.4 (9.1) & 47.6 (7.0) & 2.4 \\
offline & independent & hMup & 25.5 & 22.5 & 19.8 & 42.5 (2.6) & 59.1 (1.6) & 2.2 \\
\toprule
\end{tabular}
\end{center}
\caption{\small Comparison of mixture models with different design choices on the WMT'17 En-De dataset with $K=3$ mixture components. See \textsection\ref{subsec:model_variants}, \textsection\ref{subsec:parameterization}, \textsection\ref{subsec:schedule} for a detailed discussion about these model configurations. Pairwise-BLEU versus BLEU are plotted in Figure~\ref{fig:moe_variants}. Each configuration was run five times with different random seeds. We report the mean value of each metric (and for Pairwise-BLEU and BLEU also the standard deviation in parentheses). We also report corpus level BLEU w.r.t. one reference when greedily decoding the test set using a fixed latent variable state (columns labeled with $z=1,2,3$). Configurations that have values colored in red exhibit degeneracy of type D1, while configurations that have values colored in blue exhibit degeneracy of type D2.
}\label{tab:giant_table}
\end{table*}

\end{document}